\documentclass[15pt,conference]{IEEEtran}
\IEEEoverridecommandlockouts
\DeclareUnicodeCharacter{00A0}{ }
\usepackage{cite}
\usepackage{amsmath,amssymb,amsfonts}
\usepackage{algorithmic}
\usepackage{graphicx}
\usepackage{textcomp}
\usepackage{xcolor}
\def\BibTeX{{\rm B\kern-.05em{\sc i\kern-.025em b}\kern-.08em
    T\kern-.1667em\lower.7ex\hbox{E}\kern-.125emX}}
\begin{document}

\title{Interpretable Underwater Diver Gesture Recognition\\

}

\author{\IEEEauthorblockN{Sudeep Mangalvedhekar}
\IEEEauthorblockA{\textit{Department of Information Technology} \\
\textit{Pune Institute Of Computer Technology}\\
Pune, India \\
sudeepm117@gmail.com}
\and
\IEEEauthorblockN{Shreyas Nahar}
\IEEEauthorblockA{\textit{Department of Information Technology} \\
\textit{Pune Institute Of Computer Technology}\\
Pune, India \\
shreyashnahar0@gmail.com}
\and
\IEEEauthorblockN{Sudarshan Maskare}
\IEEEauthorblockA{\textit{Department of Information Technology} \\
\textit{Pune Institute Of Computer Technology}\\
Pune, India \\
sudarshanmaskare@gmail.com}
\and
\IEEEauthorblockN{Kaushal Mahajan}
\IEEEauthorblockA{\textit{Department of Information Technology} \\
\textit{Pune Institute Of Computer Technology}\\
Pune, India  \\
kaushalmahajan08@gmail.com}
\and
\IEEEauthorblockN{Dr Anant Bagade}
\IEEEauthorblockA{\textit{Department of Information Technology} \\
\textit{Pune Institute Of Computer Technology}\\
Pune, India  \\
ambagade@pict.edu}

}

\maketitle

\begin{abstract}
In recent years, usage and applications of Autonomous Underwater Vehicles has grown rapidly. Interaction of divers with the AUVs remains an integral part of the usage of AUVs for various applications and makes building robust and efficient underwater gesture recognition systems extremely important. In this paper, we propose an Underwater Gesture Recognition system trained on the Cognitive Autonomous Diving Buddy Underwater gesture dataset using deep learning that achieves 98.01\% accuracy on the dataset, which to the best of our knowledge is the best performance achieved on this dataset at the time of writing this paper. We also improve the Gesture Recognition System Interpretability by using XAI techniques to visualize the model's predictions.
\end{abstract}

\begin{IEEEkeywords}
Underwater Gesture recognition, Deep Learning, Machine Learning, Autonomous Underwater vehicle
\end{IEEEkeywords}

\section{Introduction}
\par In recent years, the usage of Underwater Autonomous Vehicles has grown rapidly and AUVs are used in a vast variety of applications \cite{paper1}. It includes a variety of marine applications including deep sea exploration, mining, and defense applications. With such an increase in the usage of AUVs, communication of the vehicle with the diver in underwater environments in real time proves to be extremely important. As such, building systems that can detect and understand gestures is necessary and important. Computer vision and Deep learning play an important role in solving this gesture recognition task. 
\par In this paper we propose a gesture recognition system that uses Deep Learning and Convolutional neural network trained on the CADDY dataset \cite{paper4}. We use a recorded video of divers making gestures underwater in real time while collecting data for the CADDY dataset to test our gesture recognition system.
\par Model interpretability plays a crucial role in building trust in Deep Learning systems. Hence to that end, we use two XAI methods, namely, Integrated Gradients introduced in \cite{paper12} and Occlusion Sensitivity introduced in \cite{paper11} to visualize the system's predictions.
\par Using ResNet-18 architecture introduced in \cite{paper8}, we achieve an accuracy of 98.01\% on the CADDY dataset.
\section{Related Work}
\par Classical Machine Learning Algorithms and Deep Learning Models have
been used to develop gesture recognition systems that use the CADDY Underwater dataset \cite{paper4}.
\par \cite{paper5} used a Tree-based hierarchical gesture recognition system that used a Convolutional Neural Network as a backbone. Different CNNs were used as backbones including AlexNet, ResNet, and VggNet. Standalone Convolutional Neural Network was used in \cite{paper6}, and ResNet50 was used to train the system on the CADDY dataset. \cite{paper7} used Mask R-CNN as the main model that is responsible for the detection and recognition of divers.
\par Classical machine learning and computer vision techniques such as Histogram of Gradients and Visual bag of words were used in \cite{paper6} to classify diver gestures in the CADDY dataset.
\par The performance and the techniques used by aforementioned authors and projects are summarized in Table \ref{Table:1}.
\begin{center}
\begin{table}[h]
\caption{Performance of Systems on CADDY Dataset}
\centering
    \begin{tabular}{ |p{0.04\linewidth} | p{0.2\linewidth} | p{0.3\linewidth} | p{0.15\linewidth} | p{0.11\linewidth}|} 
     \hline
     Sr No. & Methodology & Model-Algorithm & Performance & Reference \\ 
     \hline
     1 & Deep Learning & Hierarchical Tree Classifier with AlexNet backbone & 95.93 \% & \cite{paper5}\\
     \hline
     2 & Deep Learning & Hierarchical Tree Classifier with ResNet backbone & 89.47 \% & \cite{paper5}\\
     \hline
     3 & Deep Learning & Hierarchical Tree Classifier with VggNet backbone & 95.87 \% & \cite{paper5}\\
     \hline
     4 & Machine Learning & Histogram of Gradients & 84.53 \% & \cite{paper6}\\
     \hline
     5 & Machine Learning & SIFT + Bag of Visual Words & 64.03 \% & \cite{paper6}\\
     \hline
     6 & Deep Learning & ResNet50 & 97.06 \% & \cite{paper6}\\
     \hline
     7 & Deep Learning & Mask R-CNN  & 0.84 mAP & \cite{paper7} \\
     
     \hline
         
    \end{tabular}

    \label{Table:1}
\end{table}
\end{center}
\section{Data}
\par
\begin{figure}[h]
    \centering
    \includegraphics[scale=0.55]{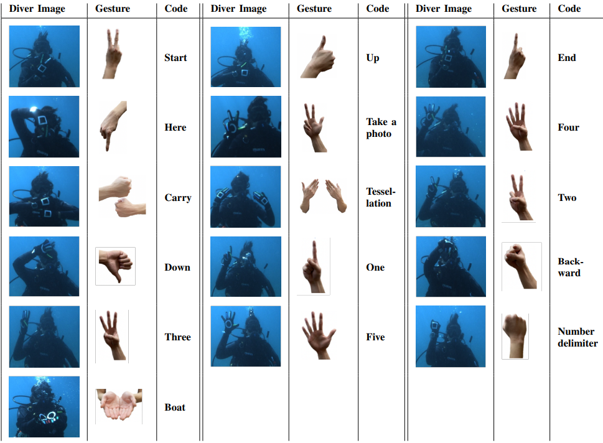}
    \caption{Gesture Classes in the CADDY dataset}
    \label{Figure:1}
\end{figure}

 Large, open source, and publicly available Datasets for Underwater diver gesture detection and recognition tasks were not available until 2018. This is when the Caddy Underwater Stereo Vision Dataset was released \cite{paper4}. The dataset was collected and maintained under the EU FP7 project. The data was collected using the BUDDY AUV developed by the Zagreb University \cite{paper3}. The environments chosen for data collection of divers making gestures underwater included open sea, indoor pools and outdoor pools. These were located at Biograd na Maru (Croatia), Geneva (Italy), and Brodarski Institute in Zagreb (Croatia). This variety of data collection environments ensures that different underwater situations are taken into consideration. The data collected in these three different scenarios was divided into eight different subgroups that represent the various diver missions and ground experiments carried out. The categories include Biograd-A, Biograd-B, and Geneva-A, which represent the trails that were done for data collection purposes and as a result include a large number of samples. Other categories include Biograd-C and Brodarski A-D which were undertaken for experimental or real diver missions. The CADDY dataset includes 17 classes or labels, with 16 classes representing various gestures made by the divers and a negative no gesture class as shown in Figure \ref{Figure:1}.
\par
\begin{figure}[h]
    \centering
    \includegraphics[scale=0.8]{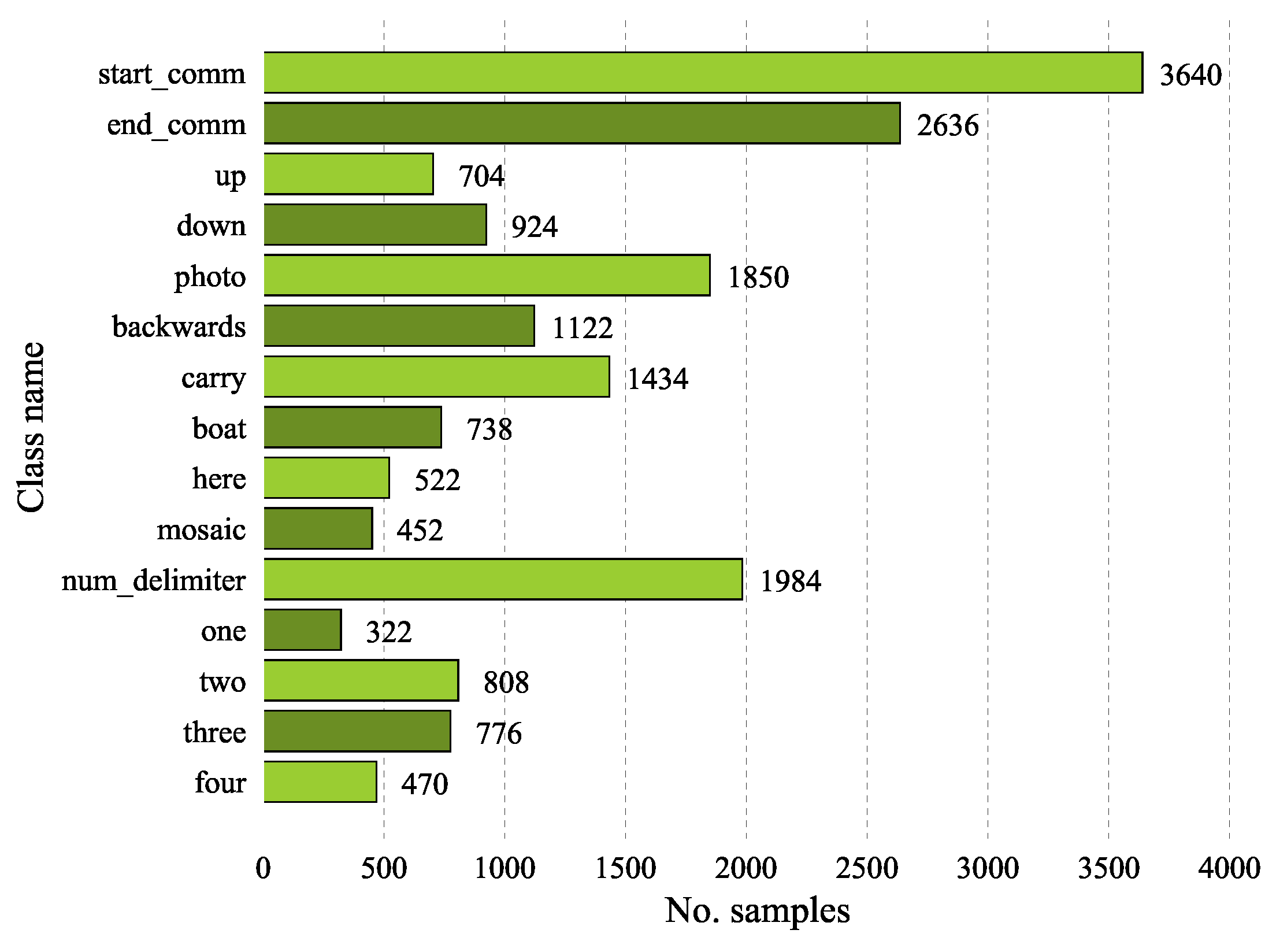}
    \caption{Class distribution in CADDY dataset}
    \label{Figure:2}
\end{figure}

 Figure \ref{Figure:2} shows a number of samples that are collected for each of the sixteen gesture classes.

\begin{figure}[h]
    \centering
    \includegraphics[scale=0.75]{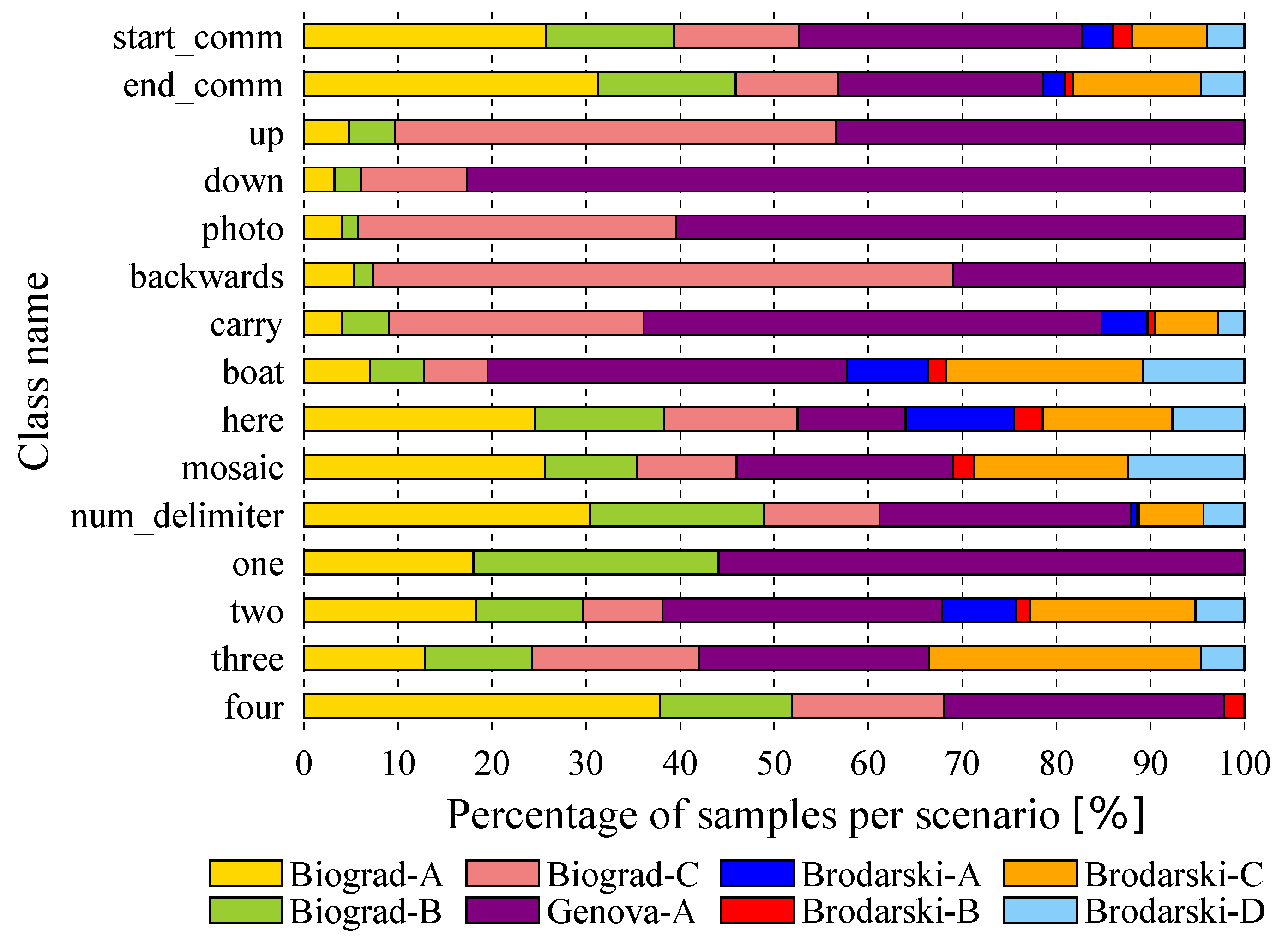}
    \caption{Class distribution per scenario in CADDY dataset}
    \label{Figure:3}
\end{figure}
\par Figure \ref{Figure:3} shows the number of samples and their distribution among the aforementioned categories under which the data collected was divided.

\section{Model Architecture}
\par For the backbone of the gesture recognition system, we used a Convolutional Neural Network for feature extraction and gesture recognition.
\par
\begin{figure}[h]
    \centering
    \includegraphics[scale=0.5]{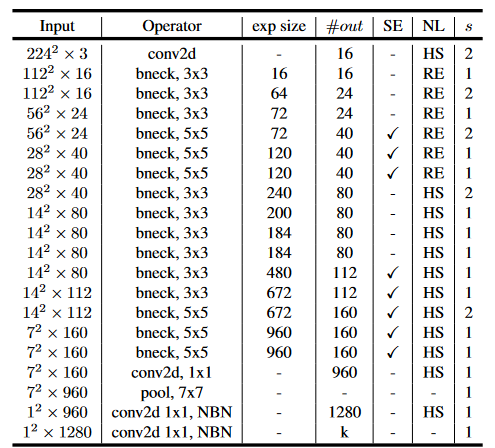}
    \caption{Mobile Net Architecture}
    \label{Figure:4}
\end{figure}

 We used MobileNet architecture introduced in \cite{paper9} and MobileNetV3 architecture introduced in \cite{paper10}. MobileNet architecture focuses on having Convolutional Neural Networks that are capable of giving high accuracy performance, yet at the same time using minimal computational resources and having efficient inference speeds. MobileNetV3 architecture as introduced in \cite{paper10} is shown in Figure \ref{Figure:4}. The notation mentioned in the figure includes, SE representing if there exists a Sequeeze and Excite block, NL denotes the type of non-linearity used and \emph{s} denotes the stride.
\par We also used ResNet Architecture introduced in \cite{paper8}. The Architecture for the ResNet block and the different architecture of a ResNet based on the depth of the network are shown in Figure 5 and Figure 6 respectively.
\begin{figure}[h]
    \centering
    \includegraphics[scale=0.65]{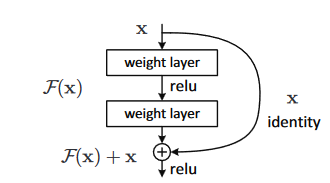}
    \caption{ResNet block}
    \label{Figure:5}
\end{figure}
\begin{figure}[h]
    \centering
    \includegraphics[scale=0.35]{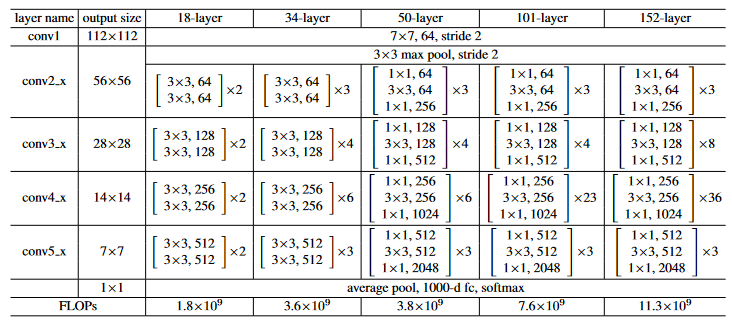}
    \caption{ResNet Architecture}
    \label{Figure:6}
\end{figure}
\\\\ \par ResNet architecture introduces a skip connection as shown in \ref{Figure:5}. This enables the construction of Deep Convolutional Neural Networks without incurring a loss of performance as mentioned in \cite{paper8}. Skip connections allow each block of a ResNet to learn a delta based on the input it receives and enable unnecessary blocks to directly learn the Identity function enabled by the connection. This reduces overfitting and improves performance as stated in \cite{paper8}.
\par We experimented with 3 different architectures namely, MobileNetV3, ResNet 50 and RestNet18 models for the backbone of our gesture recognition system. The experiments and results are discussed in Section 4.
\section{Model Interpretability}
\par Model Interpretability is extremely important to build the trust of a user in the gesture recognition system. We therefore use two model interpretability algorithms to visualize and explain the model's prediction. We used Tensorflow to implement the algorithms on our input data from the CADDY dataset. They are discussed in the following subsections.
\subsection{Integrated Gradients}
\par
\begin{figure}[h]
    \centering
    \includegraphics[scale=0.7]{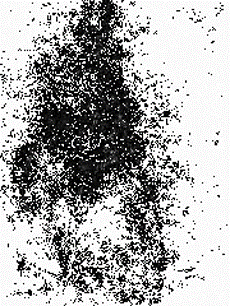}
    \caption{Integrated Gradients}
    \label{Figure:7}
\end{figure}

 Integrated Gradients as introduced in \cite{paper12} is used to explain the areas of the input, in our case an image of the diver performing a gesture, that the model uses to make the eventual prediction. It aims to establish a relationship between the features of the input image and the predictions or labels generated by the model.
Figure \ref{Figure:7} highlights the pixels used by the model to identify the gesture made for a sample input from the CADDY dataset. The black dots indicate the pixel was taken into consideration by the model while making the eventual prediction.
\subsection{Occlusion Sensitivity}
\par
\begin{figure}[h]
    \centering
    \includegraphics[scale=0.7]{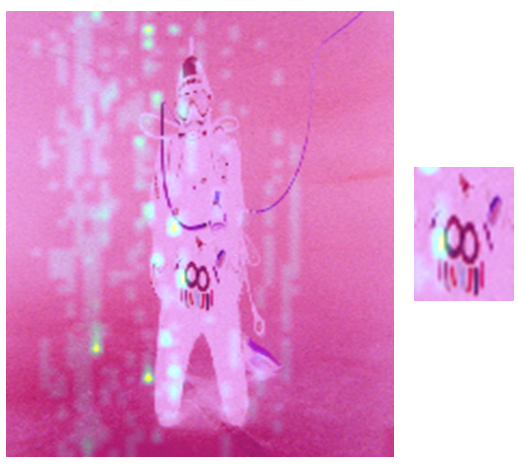}
    \caption{Occlusion Sensitivity}
    \label{Figure:8}
\end{figure}

 Occlusion Sensitivity as introduced in \cite{paper11}, is a model explainability method that aims at not only identifying the pixels or portions of the image used for prediction by also the importance of these pixels or portions on the classification of the image. Figure \ref{Figure:8} shows the occlusion sensitivity algorithm used on a sample CADDY dataset image.
\par The darker areas of the image as shown in Figure \ref{Figure:8} indicate those regions have higher importance and influence on the model's prediction of the gesture. As is expected the fingers indicate the gesture made by the diver, and as a result, as shown in Figure \ref{Figure:8}, the portion of the image corresponding to the fingers of the diver is highlighted by a darker shade.
\section{Video Classification}
\par We use a pre-recorded video of divers making gestures during the data collection phase for the CADDY dataset \cite{paper4}. We test our trained model on this video by doing a frame by frame classification.
\par Frame by frame classification of a video however introduces a flickering effect. We therefore employ a Rolling average technique to make a classification of each frame in the Video. Rolling average takes into account not only the predictions made by the model on the current frame but also on a predefined number of previous frames. The results of the same are discussed in Section 4.
\section{Experimentation and Model Training}
\subsection{Model Training and Hyperparameters}
\par We trained three different Convolutional Neural Network Architectures, namely MobileNetV3, ResNet50 and ResNet18. We used TensorFlow and Pytorch for training and evaluating the models and Google Colab for training the models on the cloud GPU.
\subsubsection{Hyperparameters}
\par We used Data augmentation techniques to improve the model robustness and performance. Image Rotation (\(0^0\) to \(20^0\)), Zoom In and Out with ratio 0.9 and 1.1 respectively and apply Normalization.
\par We used \emph{Pytorch's torch.optim.lr\_scheduler.StepLR} to decrease the learning rate by \(\sqrt{0.1}\) every 7 epochs with the base learning rate being 0.001. We use the Adam Optimizer to train the model.
\par We tried different batch size ranges from 16, 32, and 64. Models were trained for 25 - 30 epochs each.
\subsubsection{Training}
\par We used Transfer Learning for training our models for the gesture recognition task. All models were initialized with IMAGENET weights. We used the following two transfer learning approaches for transfer learning:
\begin{itemize}
    \item Feature-Extraction: Freeze the convolutional layers of the network and train only the dense / fully connected layers.
    \item Fine-Tuning: Initialize the model with pre-trained weights, such as IMAGENET weights and re-train the convolutional and dense layers of the model on the CADDY dataset.
\end{itemize}
\par Table \ref{Table:2} summarizes the model architecture and the type of transfer learning approach used:\\\\\\
\begin{center}
\begin{table}[h]
    \caption{Transfer Learning Approaches Used}
\centering
    \begin{tabular}{ |p{0.05\linewidth} | p{0.25\linewidth} | p{0.4\linewidth} | p{0.15\linewidth} | p{0.1\linewidth}|} 
     \hline
     Sr No. & Model Architecture & Approach Used & Layers Trained & Epochs \\ 
     \hline
     1 & MobileNetV3 & Feature Extraction & Dense & 40\\
     \hline
     2 & ResNet50 & Feature Extraction & Dense & 40\\
     \hline
     3 & ResNet50 & Fine Tuning & Convolutional (Partial) + Dense & 40\\
     \hline
     4 & ResNet18 & Feature Extraction & Dense & 30\\
     \hline
     5 & ResNet18 & Fine Tuning & All & 30\\
     
     \hline
         
    \end{tabular}

    \label{Table:2}
\end{table}
\end{center}
\par All models were trained on Google Colab's Cloud GPU with 16GB RAM.
\subsection{Recorded Video Classification}
\par
\begin{figure}[h]
    \centering
    \includegraphics[scale=0.4]{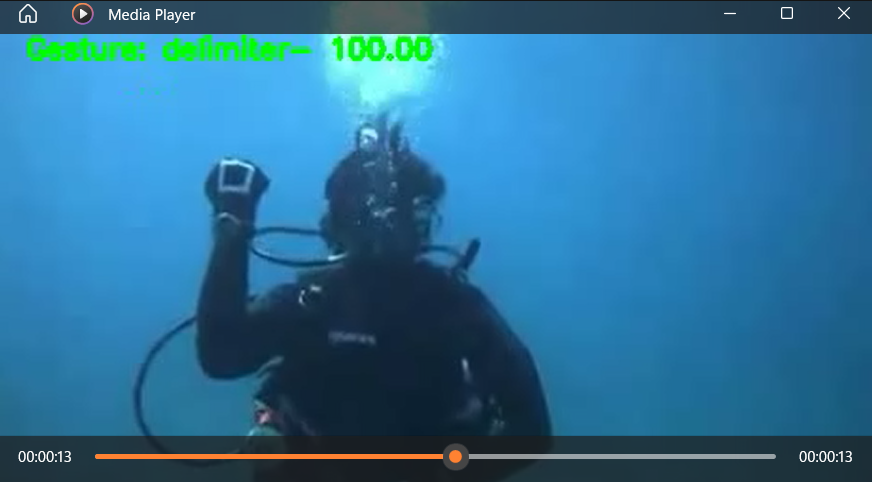}
    \caption{Frame in the recorded video}
    \label{Figure:9}
\end{figure}

 We employ a frame by frame video classification technique that uses a rolling average. Rolling average ensures flickering effect is not introduced, which leads to rapid changes between prediction between consecutive frames, and instead leads to a smooth frame by frame prediction. Figure \ref{Figure:9} shows one of the frames from the video that are classified correctly by the model.
\par Each frame in the video is classified using the gesture recognition model trained on the CADDY dataset. Each frame is annotated with the gesture class and the confidence of the system in classifying the gesture represented in percentage, \emph{delimiter} and 100 in Figure \ref{Figure:9} respectively. Video is processed using the OpenCV library. 

\section{Results}
\subsection{Model Accuracy}
\par We achieve the best performance with the ResNet18 model trained for 30 epochs with a batch size of 64, with a test accuracy of 98\%. Table 3 summarizes the accuracies for each of the models trained.
\begin{center}
\begin{table}[h]
    \caption{Model Performance}
\centering
    \begin{tabular}{ |p{0.05\linewidth} | p{0.25\linewidth} | p{0.15\linewidth} | p{0.15\linewidth}|} 
     \hline
     Sr No. & Model Architecture & Epochs & Accuracy \\ 
     \hline
     1 & MobileNetV3 & 40 & 84.32 \%\\
     \hline
     2 & ResNet50 & 40 & 92.3 \%\\
     \hline
     3 & ResNet18 & 30 & 98 \%\\
     \hline
         
    \end{tabular}

    \label{Table:3}
\end{table}
\end{center}
\par The test set contained 3093 images from the CADDY dataset and the accuracy metric was used to evaluate performance.
\subsection{Model Confidence Analysis}
\par We test our best performing model, based on the ResNet18 architecture by using the confidence scores resulting from the Softmax layer. Table \ref{Table:3} shows the confidence scores for all 17 gesture classes.
\begin{center}
\begin{table}[h!]
    \caption{Model Confidence}
\centering
    \begin{tabular}{ |p{0.05\linewidth} | p{0.25\linewidth} | p{0.15\linewidth} | p{0.15\linewidth}|} 
     \hline
     Sr No. & Gesture Class & Confidence\\ 
     \hline
     1 & Backward &  98.090 \%\\
     \hline
     2 & Boat &  99.560 \%\\
     \hline
     3 & Carry &  99.604 \%\\
     \hline
     4 & Delimiter &  99.75 \%\\
     \hline
     5 & Down &  99.955 \%\\
     \hline
     6 & End &  99.418 \%\\
     \hline
    7 & Five &  98.073 \%\\
     \hline
    8 & Four &  99.805 \%\\
     \hline
    9 & Here &  99.253 \%\\
     \hline
    10 & Mosaic &  99.999 \%\\
     \hline
    11 & None &  99.072 \%\\
     \hline
    12 & One &  99.902 \%\\
     \hline
    13 & Photo &  99.213 \%\\
     \hline
    14 & Start &  99.672 \%\\
     \hline
    15 & Three &  99.542 \%\\
     \hline
    16 & Two &  99.215 \%\\
     \hline
    17 & Up &  98.709 \%\\
     \hline
    \hline
    \end{tabular}

    \label{Table:4}
\end{table}
\end{center}
\par High confidence scores for all gesture classes indicate the accurate and robust nature of the gesture recognition system trained using the ResNet18 backbone.

\newpage
\section{Conclusion and Future Work}
\par Communication between the diver and the robot is of utmost importance for effective usage of Autonomous Underwater Vehicles. This project aims at solving a part of that by building an Underwater Gesture Recognition System. 
\par We have implemented a gesture recognition system using a deep learning model for identifying gestures within the Caddyian language. Our model architecture is based on ResNet18 and achieved a test accuracy of 98\%. It is as per our knowledge at the time of this paper the best performing model in terms of test accuracy on the Caddy underwater gestures dataset.
We further implement a video processing pipeline that uses a Rolling average technique to predict the gestures in the video feed in real time. XAI algorithms such as Integrated Gradients and Occlusion sensitivity are implemented to produce visualizations for model Interpretability.
\par Based on the error analysis performed future work will investigate, Generative Adversarial Networks (GAN) to generate more samples for the CADDIYAN gesture images for training the network, particularly those that have a limited number of samples. A more complex CNN can be used which includes a binary classifier to classify the true negative samples followed by a CNN to classify the other gesture classes.

\vspace{12pt}

\end{document}